\pdfoutput=1

\documentclass[11pt]{article}

\usepackage{emnlp2021}

\usepackage{times}
\usepackage{latexsym}
\usepackage{iitem}
\usepackage[utf8]{inputenc}
\usepackage{microtype}
\usepackage{graphicx}
\usepackage{subfigure}
\usepackage{booktabs} 
\usepackage{comment}
\usepackage{colortbl}
\usepackage{arydshln}
\usepackage{makecell}
\usepackage{hyperref}
\usepackage{mathtools}
\usepackage{color}
\usepackage{graphicx}
\usepackage{CJKutf8}
\usepackage{url}
\usepackage{multicol,multirow}
\usepackage{makecell}
\usepackage{array}
\usepackage{bm}
\usepackage{pifont}
\usepackage{parskip}
\usepackage[T1]{fontenc}    
\usepackage{booktabs}       
\usepackage{amsmath, amsthm}
\usepackage{amssymb}
\usepackage{amsfonts}

\usepackage{enumitem}
\usepackage{microtype}

\newcommand{\heart}{\text{\small \ding{170}}}

\title{$k$Folden: $k$-Fold Ensemble for Out-Of-Distribution Detection}

\author{
Xiaoya Li$^\spadesuit$, 
Jiwei Li$^{\clubsuit\spadesuit}$, 
Xiaofei Sun$^\spadesuit$,
Chun Fan$^{\bigstar\blacktriangle\blacktriangledown}$\\
{\bf 
Tianwei Zhang$^{\blacklozenge}$,
Fei Wu$^{\clubsuit}$,
Yuxian Meng$^\spadesuit$,
Jun Zhang$^\heart$
}\\ 
$^\spadesuit$Shannon.AI\\
$^\blacktriangledown$National Biomedical Imaging Center, Peking University\\
$^\bigstar$Computer Center of Peking University, 
$^\blacktriangle$Peng Cheng Laboratory \\
$^\blacklozenge$Nanyang Technological University
$^\clubsuit$Zhejiang University, $^\heart$Tsinghua University\\ 
\{xiaoya\_li, jiwei\_li, xiaofei\_sun, yuxian\_meng\}@shannonai.com\\
fanchun@pku.edu.cn, tianwei.zhang@ntu.edu.sg\\
wufei@zju.edu.cn, jun-zhan19@mails.tsinghua.edu.cn
}

\begin{document}
\maketitle
\begin{abstract}
Out-of-Distribution (OOD) detection is an important problem in natural language processing (NLP). In this work, we propose a simple yet effective framework $k$Folden, which mimics the behaviors of OOD detection during training without
the use of
any external data. For a task with $k$ training labels, $k$Folden induces $k$ sub-models, each of which is trained on a subset with $k-1$ categories with the left category masked unknown to the sub-model. Exposing an unknown label to the sub-model during training,
the model is encouraged to learn
to equally attribute the 
 probability  to the seen $k-1$ labels for the unknown label, enabling  this framework to simultaneously resolve in- and out-distribution examples in a natural way via OOD simulations.
Taking text classification as an archetype, we develop benchmarks for OOD detection using existing text classification datasets. By conducting comprehensive comparisons and analyses on the developed benchmarks, we demonstrate the superiority of $k$Folden against current methods in terms of improving OOD detection performances while maintaining improved in-domain classification accuracy. The code and datasets can be found at: \url{https://github.com/ShannonAI/kfolden-ood-detection}. \footnote{Corresponding author: Jun Zhang.}
\end{abstract}

\section{Introduction}
\label{introduction}
Recent progress in deep neural networks has drastically improved accuracy in numerous NLP tasks \cite{sun2019fine,raffel2019exploring,chai2020description,he2020deberta}, but detecting out-of-distribution (OOD) examples from the in-domain (ID) examples is still a challenge for existing state-of-the-art deep NLP models. The ability of identifying OOD examples is critical for building reliable and trustworthy NLP systems for, say, text classification \cite{hendrycks2016baseline,mukherjee2020uncertainty}, question answering \cite{kamath2020selective} and neural machine translation \cite{kumar2019calibration}. 
Existing works studying OOD detection in NLP  often rely on external data \cite{hendrycks2018deep} to diversify model predictions and achieve better generality in OOD detection. 
The reliance on external data not only brings additional burden for data collection, but also results in the annoying issue in deciding which subset of external data to use: there is massive amount of external data and the using different subsets  leads to different final results.
Therefore, developing 
OOD detection system 
without external data is important towards building reliable NLP systems.

In this work, we propose a novel, simple yet effective framework, $k$Folden, short for a $\bm{k}$-{\bf Fold} {\bf en}semble, to address OOD detection for NLP without the use of any external data. We accomplish this goal by simulating the process of detecting OOD examples {\it during training}. 
Concretely, for a standard NLP task with $k$ labels for both training and test, 
we first obtain $k$ separate sub-models, each of which is trained on a set of different $k-1$ labels with the left one being masked unknown to the model. We train each sub-model by jointly optimizing the cross entropy loss for the visible $k-1$ labels and the KL divergence loss between the predicted distribution and the uniform distribution for the left-one-out label. 
During test, we simply average the probability distributions produced by these $k$ sub-modules and treat the result as the final probability estimate for a given input. Intuitively, if the input is an ID example, the final probability distribution will lay much of the weight on one of the $k$ seen labels, but if the input is an OOD example, we expect the final probability distribution to get close to the uniform distribution, since each sub-model has tried to even its probability distribution when encountering unseen labels during training. 

This training paradigm does not rely on any external data, and by mimicking the behaviors of distinguishing unseen labels from the seen, i.e., simulating the process of OOD detection during training, which is completed via the KL divergence loss, the framework naturally detects OOD examples and is able to perform reasonably better than other widely used strong OOD detection methods. Moreover, $k$Folden is complementary to existing post-hoc OOD detection methods, and combining both leads to the most performance boosts.

To facilitate OOD detection researches in NLP, we also construct  benchmarks on top of four widely used text classification datasets: 20NewsGroups, Reuters, AG News and Yahoo!Answers. This created benchmark consists of 7 datasets with different levels of difficulty directed to two types of OOD examples: semantic shift and non-semantic shift, which differ in whether a shift is related to the inclusion of new semantic categories. The proposes benchmarks help comprehensively examine OOD detection methods, and we hope it can serve as a convenient and general tool for developing more robust and effective OOD detection models.

To summarize, the contributions of this work are:
\begin{itemize}[noitemsep,topsep=0pt,parsep=0pt,partopsep=0pt]
    \item We propose a simple yet effective framework -- $k$Folden, which simulates the process of OOD detection during training without using any external data. 
    \item We construct benchmarks for  OOD detection in text classification hoping for facilitating future related researches.
    \item We conduct comprehensive comparisons and analyses between existing methods and the proposed $k$Folden on the benchmark, and we show that $k$Folden achieves  performance boosts regarding OOD detection while maintaining ID classification accuracy. 
  \end{itemize}

\section{Related Work}
\label{related}

\paragraph{Out-Of-Distribution Detection}~{}\\
Detecting OOD examples using deep neural models has gained substantial traction over recent years.
\citet{hendrycks2016baseline} proposed a baseline for misclassified and OOD examples by thresholding candidates based on the predicted softmax class probability. 
\citet{lee2018simple} trained a classifier concurrent with a generator under the GAN framework \cite{goodfellow2014generative}. The generator produces examples at the in-domain boundary and the classifier is forced to give lower confidence in predicting the classes for those examples. \citet{hendrycks2018deep} leveraged real datasets instead of the generated examples, enabling the classifier to better generalize and detect anomalies.
\citet{liang2017enhancing} observed that temperature scaling and small perturbations lead to widened gaps between ID and OOD examples, for which they proposed proposed ODIN, a technique that makes OOD instances distinguishable by pulling apart the softmax scores of ID and OOD examples.
\citet{kamath2020selective} proposed to leverage the confidence estimate of a QA model to determine whether a question should be answered under domain shift to maintain a moderate accuracy. 
\citet{hendrycks2019using,hendrycks2020pretrained} showed that pretraining improves model robustness in terms of uncertainty estimation and OOD detection.
Measuring model confidence has also exhibited power in detecting OOD examples \cite{lee2017confident,lee2017training,devries2018learning,papadopoulos2021outlier}.
This work differs from \citet{hendrycks2020pretrained} mainly in that (1) they used a simple MaxProb-based method \cite{hendrycks2016baseline} to estimate uncertainty while we propose a novel framework $k$Folden to improve OOD detection; and (2) they focused on comparing different NLP models on OOD generalization and shed light on the importance of pretraining for OOD robustness, whereas we highlight the merits of OOD simulation during training without the use of any external data, and construct a dedicated benchmark for text classification OOD detection.

\paragraph{Meta Learning in NLP}~{}\\
Meta learning \cite{thrun2012learning,andrychowicz2016learning,nichol2018first,finn2017model} tackles the problem of model learning in the domain with scarce data when large quantities of data are accessible in another related domain.
Meta learning has been applied to considerable NLP tasks including semantic parsing \cite{huang2018natural,guo2019coupling,sun2020neural}, dialog generation \cite{song2019learning,huang2020meta}, text classification \cite{wu2019learning,sun2020neural,bansal2020self,lin2021bertgcn} and machine translation \cite{gu2018meta}.
Our work is distantly related to meta learning in terms of the way we train $k$Folden by simulating the behaviors of predicting the unseen label during training, But we do not intend to achieve strong few-shot learning performances, which is the main goal of meta learning.

\section{Task Definition}
\label{task}
In this paper, we consider the problem of distinguishing betweem ID and OOD examples. We take text classification for illustration, and other tasks can be analogously resolved using the proposed $k$Folden framework. Let $\mathcal{D}^\text{train}=\{\bm{x},y^\text{train}\}$ and $\mathcal{D}^\text{test}=\{\bm{x},y^\text{test}\}$ denote the two sets respectively used for model training and test, where we assume the label space for training consists of $k$ distinct labels $\mathcal{Y}^\text{train}=\{1,\cdots,k\}$ and all possible labels for test is the ones in $\mathcal{Y}^\text{train}$ plus $t$ additional labels, i.e., $\mathcal{Y}^\text{test}=\{1,\cdots,k,k+1,\cdots,k+t\}$. Assume that a neural network $f$ is trained on $\mathcal{D}^\text{train}$, and tested on $\mathcal{D}^\text{test}$.

We are interested in two situations when testing $f$ on $\mathcal{D}^\text{test}$: (1) the current input example $\bm{x}$ has a gold label belonging to $\mathcal{Y}^\text{train}$ (i.e., $y^\text{test}\in\mathcal{Y}^\text{train}$), and (2) the input example's gold label does not belong to $\mathcal{Y}^\text{train}$ (i.e., $y^\text{test}\in\mathcal{Y}^\text{test}\backslash\mathcal{Y}^\text{train}$). For the former, we would like the model to achieve high accuracy because it has been trained on these ID examples; for the latter, we expect the model to figure out the current input is an OOD example. Hence, in this work, we mainly report the results from two aspects: accuracy on ID examples, and performances on OOD examples. The performance for OOD examples is evaluated via several targeted metrics, which will be introduced in experiments.

\section{Method: $k$-Fold Ensemble}
\label{method}

\subsection{Training $k$ Sub-Models as Simulation for OOD Detection}
The core idea behind the proposed $k$Folden framework is to simulate the situation of encountering 
 unseen labels  at the training stage without the use of external data. To this end, we propose to train $k$ independent sub-models $\{f_1,\cdots,f_k\}$, each of which is in order trained on a different subset of $k-1$ labels with the left label masked unknown to the model. Each sub-model is required to attain high accuracy on 
 examples with
 the seen $k-1$ labels along with high uncertainty on examples with of masked label, and this is exactly what we would expect for OOD detection: we would like the model to accurately detect OOD examples while not harming performances on ID examples.

More specifically, assume we are training the $i$-th sub-model $f_i 
(1\le i\le k)$, and thus the visible label set for training $f_i$ would be $\mathcal{Y}^\text{train}\backslash\{i\}$. All training examples in 
$\mathcal{D}^\text{train}$ with label $i$ now becomes unknown to $f_i$. For the visible $k-1$ labels, $f_i$ should still achieve high accuracy as we want; but for the masked label $i$, $f_i$ needs to give non-deterministic estimates when the input instance $\bm{x}$ has the ground-truth label $i$ because the label $i$ is masked and not found in the training set. 
This implies 
 that the model can not determine which label $\bm{x}$ belongs to and may attribute it to an OOD example. These two considerations can be satisfied  by jointly optimizing the following objective:
\begin{equation}
\begin{aligned}
    \mathcal{L}=\mathcal{L}_\text{CE}+ \gamma \mathcal{L}_\text{KL}
\end{aligned}
\label{eq1}
\end{equation}
where 
\begin{equation}
  \begin{aligned}
    \mathcal{L}_\text{CE}=\sum_{\substack{(\bm{x},y^\text{train})\in\mathcal{D}^\text{train}\\y^\text{train}\in\mathcal{Y}^\text{train}\backslash\{i\}}}\texttt{CrossEntropy}(y^\text{train}, f_i(\bm{x}))
  \end{aligned}
\label{eq2}
\end{equation}
\begin{equation}
  \begin{aligned}
    \mathcal{L}_\text{KL}=\sum_{\substack{(\bm{x},y^\text{train})\in\mathcal{D}^\text{train}\\y^\text{train}=i}}\texttt{KL}(f_i(\bm{x}),\bm{u})
  \end{aligned}
\label{eq3}
\end{equation}
$\gamma$ is a hyper-parameter ranging over $[0, 1]$ and tuned on validation set. 
In the above equations, $\bm{u}$ is a uniform distribution. 
Eq.(\ref{eq2}) is a standard cross entropy loss that requires the model to achieve accurate predictions on the visible labels, while Eq.(\ref{eq3}) draws on the KL divergence to encourage the model to produce a probability distribution close to the uniform distribution $\bm{u}$
on the $k-1$ labels
for the masked label. By jointly training on both loss functions, $f_i$ will be able to detect the OOD label $i$ while preserving non-reduced performances on other $k-1$ labels.
We proceed with this process for all $k$ sub-models, each with a different masked label.
$f_i(\bm{x})$ takes as input $\bm{x}$ and outputs a probability distribution of dimensionality $k-1$. $f_i$ can be implemented using any model backbone such as LSTM \cite{hochreiter1997long}, CNN \cite{kim-2014-convolutional}, Transformer \cite{vaswani2017transformer} and BERT \cite{devlin2018bert}.

\subsection{Sub-Model Ensemble}
\label{sec:inference}
A single sub-model $f_i$ will inevitably result in poor performances during test regarding the ID examples with label $i$. This is because for $f_i$, the masked label $i$ during training will never have the chance to be predicted by the model, so that all the test examples with label $i$ in $\mathcal{D}^\text{test}$ will be associated with possibly low probability, leading to overall reduced accuracy. 

To tackle this issue, we adopt the idea of model ensemble: given an input $\bm{x}$, we first obtain $k$ probability distributions $\{f_1(\bm{x}),\cdots,f_k(\bm{x})\}$ respectively produced by the $k$ sub-models. In order to coordinate the label dimensions for different sub-models, we manually pad a zero dimension to each probability distribution at the corresponding masked position. For example, if $k=4$ and the output from $f_2$ is $f_2(\bm{x})=[f_2(\bm{x})_1,f_2(\bm{x})_2,f_2(\bm{x})_3,f_2(\bm{x})_4]$, then the padded output distribution would thus be $\tilde{f}_2(\bm{x})=[f_2(\bm{x})_1,0,f_2(\bm{x})_2,f_2(\bm{x})_3,f_2(\bm{x})_4]$. Next, we average all the $k$ padded probability distributions, and take the result as the final probability estimate:
\begin{equation}
  \begin{aligned}
    \tilde{f}(\bm{x})=\frac{1}{k}\sum_{i=1}^k\tilde{f}_i(\bm{x})
  \end{aligned}
\label{eq4}
\end{equation}
$\tilde{f}(\bm{x})$ is still a valid probability distribution and naturally remedies the shortcoming of a single sub-model: if $\bm{x}$ is an ID example, i.e., its ground-truth label $y$ belongs to $\mathcal{Y}^\text{train}$, $\tilde{f}(\bm{x})$ will put most of the probability mass on one of the $k$ labels; if $\bm{x}$ is an OOD example, $\tilde{f}(\bm{x})$ will get close to the uniform distribution because all sub-models comprising $\tilde{f}(\bm{x})$ will even their probability masses across all the $k$ labels. After training, $\tilde{f}(\bm{x})$ can be used for ID evaluation and OOD evaluation
simultaneously.

\begin{table*}[t]
  \centering
  \small
  \scalebox{0.7}{
  \begin{tabular}{lccccccc}\toprule
    & \multicolumn{3}{c}{{\it Non-Semantic Shift (NSS) Datasets}} & \multicolumn{4}{c}{{\it Semantic Shift (SS) Datasets}}\\
    \cmidrule(r){2-4}\cmidrule(r){5-8}\\
    & {\bf 20News-6S} & {\bf AG-EXT} & {\bf Yahoo-AG-five} & {\bf Reuters-$m$K-$n$L} & {\bf AG-FL} & {\bf AG-FM} & {\bf Yahoo-FM}\\\midrule
    {\it Adapted From} & 20News & AGNews\&AGCorpus & Yahoo\&AGCorpus & Reuters & AGNews\&AGCorpus & AGNews\&AGCorpus & Yahoo \\ 
    \specialrule{0em}{1pt}{1pt}
    \cdashline{1-8}
    \specialrule{0em}{1pt}{1pt}
    {\it \# Labels in T} & 6 & 4 & 5 & $m$ & 4 & 4 & 5 \\ 
    {\it \# Instances in T} & 5,600 & 112,400 & 675,000 & $f(m,\text{train})$ & 116,000 & 116,000 & 680,000 \\ 
    \specialrule{0em}{1pt}{1pt}
    \cdashline{1-8}
    \specialrule{0em}{1pt}{1pt}
    {\it \# Labels in ID-V} & 6 & 4 & 5 & $m$ & 4 & 4 & 5 \\ 
    {\it \# Instances in ID-V} & 800 & 7,600 & 25,000 & $f(m,\text{valid})$ & 4,000 & 4,000 & 20,000 \\ 
    {\it \# Labels in OOD-V} & 6 & 4 & 5 & $n$ & 4 & 4 & 5 \\ 
    {\it \# Instances in OOD-V} & 800 & 7,600 & 25,000 & $f(n,\text{valid})$ & 4,000 & 4,000 & 20,000 \\ 
    \specialrule{0em}{1pt}{1pt}
    \cdashline{1-8}
    \specialrule{0em}{1pt}{1pt}
    {\it \# Labels in ID-T} & 6 & 4 & 5 & $m$ & 4 & 4 & 5 \\ 
    {\it \# Instances in ID-T} & 800 & 7,600 & 25,000 & $f(m,\text{test})$ & 4,000 & 3,603 & 25,000 \\ 
    {\it \# Labels in OOD-T} & 6 & 4 & 5 & $n$ & 4 & 4 & 5 \\ 
    {\it \# Instances in OOD-T} & 800 & 7,600 & 25,000 & $f(n,\text{test})$ & 4,000 & 3,603 & 25,000 \\\bottomrule
  \end{tabular}
  }
  \caption{Statistics for the constructed benchmark. ``T'' is for ``Training Set'', ``V'' is for ``Valid Set'', and ``T'' is for ``Test Set''. All the data in each of the set are evenly distributed over the labels except 20News-6S. $f(m,\text{train/valid/test})$ means that the actual number is related to $m$ and the corresponding train/valid/test set in the original Reuters-ModApte dataset.}
  \label{tab:benchmark}
\end{table*}

\begin{table}[!t]
    \centering
    \scalebox{0.55}{
    \begin{tabular}{lcccc}\toprule
    & \multicolumn{1}{c}{\bf ID Metrics} & \multicolumn{3}{c}{\bf OOD Metrics}\\
    \cmidrule(r){2-2}\cmidrule(r){3-5}
    {\bf Model} &{\bf ACC$\uparrow$} & {\bf AUROC$\uparrow$}  & {\bf AUPR$\uparrow$}  & {\bf TNR@95TPR$\uparrow$}   \\\midrule
    \multicolumn{5}{c}{\underline{\it 20Newsgroups-6S}}\vspace{1pt}\\
    \multicolumn{5}{c}{\bf Vanilla}\vspace{1pt}\\
    CNN-init emb & 77.76  & 50.22 & 58.91 & 29.27 \\ 
    BiLSTM-init emb & 78.01 & 50.00 & 59.93 & 29.53 \\ 
    BERT & 82.15 & 54.76  & 62.61  & 50.89 \\
    RoBERTa & 83.40  & 57.41 & 66.79 & 59.15 \\
    RoBERTa+Mahalanobis & 83.40  & {58.22} & {68.63}  & {61.98} \\
    RoBERTa-Dropout & {85.06} & 57.72 & 67.30 & 60.56 \\
    RoBERTa-Dropout+Mahalanobis &  {\bf 85.06} &  58.29 & 68.99 & {\bf  62.40} \\
    RoBERTa$(6)$ &  84.37 & 58.00 & 67.94 & 60.81  \\
    RoBERTa$(6)$+Mahalanobis & 84.69 & {\bf 58.42} & {\bf 69.05} &  62.07  \\
    \specialrule{0em}{1pt}{1pt}
    \cdashline{1-5}
    \specialrule{0em}{1pt}{1pt}
    \multicolumn{5}{c}{\bf $k$Folden}\vspace{1pt}\\
    CNN-init emb & 78.29  & 50.33 & 62.10 & 34.57 \\ 
    BiLSTM-init emb & 78.30 & 50.48 & 60.86 & 34.94 \\ 
    BERT & 84.12 &  56.77 & 64.85 & 53.46 \\
    RoBERTa & {85.75} & 58.35 & 67.54 & 60.45 \\
    RoBERTa+Scaling & {85.75} & 59.83 & 68.88  & 62.17 \\
    RoBERTa+Mahalanobis & {\bf 85.75} & {\bf 60.04} & {\bf 69.91}  & {\bf 63.44}  \\\hline
    \multicolumn{5}{c}{\underline{\it AGNews-EXT}}\vspace{1pt}\\
    \multicolumn{5}{c}{\bf Vanilla}\vspace{1pt}\\
    CNN-init emb & 86.13 & 48.29 & 61.54 & 35.62  \\ 
    BiLSTM-init emb & 87.38 & 48.56  & 62.15 & 35.88 \\ 
    BERT & 92.24 & 51.35 & 63.68  & 49.63  \\
    RoBERTa & 94.54 & 52.75  & 64.01  & 51.45   \\
    RoBERTa+Mahalanobis &  94.54 & {55.37} & {65.94}  & {54.60} \\
    RoBERTa-Dropout & {95.13} & 52.74& 64.32 & 52.47   \\
    RoBERTa-Dropout+Mahalanobis & 95.13 & 55.67  & 66.32 & 55.10 \\
    RoBERTa$(4)$ & 95.22 & 53.91 & 65.68 & 53.08 \\
    RoBERTa$(4)$+Mahalanobis &  {\bf  95.22} & {\bf 55.74} & {\bf 66.58} & {\bf 55.21}  \\
    \specialrule{0em}{1pt}{1pt}
    \cdashline{1-5}
    \specialrule{0em}{1pt}{1pt}
    \multicolumn{5}{c}{\bf $k$Folden}\vspace{1pt}\\
    CNN-init emb & 88.30  & 49.31 & 62.18 & 37.20 \\ 
    BiLSTM-init emb & 88.92 & 49.45 & 63.08 & 37.54 \\ 
    BERT & 93.43  &  51.25 & 64.19 & 53.16 \\
    RoBERTa & 95.62 & 53.87  &  65.76 & 54.98 \\
    RoBERTa+Scaling & {95.62} & 55.19 & 66.28  & 55.09 \\
    RoBERTa+Mahalanobis & {\bf 95.62}  & {\bf 56.07}  & {\bf 67.81}  & {\bf 55.48} \\\hline
    \multicolumn{5}{c}{\underline{\it Yahoo-AGNews-five}}\vspace{1pt}\\
    \multicolumn{5}{c}{\bf Vanilla}\vspace{1pt}\\
    CNN-init emb & 77.45 & 79.26 & 58.50 & 43.94 \\ 
    BiLSTM-init emb & 77.68 & 79.98 & 58.76 & 44.07 \\ 
    BERT &  81.93 & 82.35 & 62.17 & 50.82  \\
    RoBERTa & 82.54 & 84.98  & 63.46 & 50.94 \\
    RoBERTa+Mahalanobis & 82.54& {85.88} & {63.92}  & {51.96}  \\
    RoBERTa-Dropout & {84.04} &  84.29 & 63.36  & 51.03 \\
    RoBERTa-Dropout+Mahalanobis & 84.04 & 85.95  & 64.17 &  51.39  \\
    RoBERTa$(5)$ &  84.10 & 85.01 & 64.14 & 51.22 \\
    RoBERTa$(5)$+Mahalanobis & {\bf 84.10} & {\bf 86.23} & {\bf 64.37} & {\bf 53.11}  \\
    \specialrule{0em}{1pt}{1pt}
    \cdashline{1-5}
    \specialrule{0em}{1pt}{1pt}
    \multicolumn{5}{c}{\bf $k$Folden}\vspace{1pt} \\
    CNN-init emb & 79.23  & 81.12 & 61.09 & 45.82  \\ 
    BiLSTM-init emb & 78.04 &  82.33 & 62.88 & 45.90 \\ 
    BERT & 83.23 &  84.09 &  63.11 & 52.95 \\
    RoBERTa & {84.45} & 85.61 & 64.15 & 52.22  \\
    RoBERTa+Scaling & {84.45} & 86.69 & 64.87  & 54.39 \\
    RoBERTa+Mahalanobis &  {\bf 84.45} & {\bf 86.92} &  {\bf 64.92} & {\bf 56.24} \\\bottomrule
    \end{tabular}
    }
    \caption{Results of Non-Semantic Shift (NSS) datasets. The number in the bracket $(k)$ denotes averaging $k$ model predictions and $k$ equals to the number of labels in the training dataset.}
    \label{exp:nss}
    \end{table}
\begin{table}[!t]
    \centering
    \scalebox{0.55}{
    \begin{tabular}{lcccc}\toprule
    & \multicolumn{1}{c}{\bf ID Metrics} & \multicolumn{3}{c}{\bf OOD Metrics}\\
    \cmidrule(r){2-2}\cmidrule(r){3-5}
    {\bf Model} &{\bf ACC$\uparrow$} & {\bf AUROC$\uparrow$}  & {\bf AUPR$\uparrow$}  & {\bf  TNR@95TPR$\uparrow$}   \\\midrule
    \multicolumn{5}{c}{\underline{\it Reuters-7K-3L}}\vspace{1pt}\\
    \multicolumn{5}{c}{\bf Vanilla}\vspace{1pt}\\
    CNN-init emb & 62.04 & 64.76 & 53.49 & 49.23 \\ 
    BiLSTM-init emb & 60.89 & 66.41  & 55.55 & 48.58 \\ 
    BERT & 63.25 &  66.83 & 60.28 & 50.66    \\
    RoBERTa & 65.88 & 67.37  & 63.30 & 51.95  \\
    RoBERTa+Mahalanobis & 65.88 & 68.34  &  {64.33} & {52.73}  \\
    RoBERTa-Dropout &  66.16 &  69.04 & 64.18  & 52.09 \\
    RoBERTa-Dropout+Mahalanobis & 66.16 & 69.86  & 64.25 & 52.90 \\
    RoBERTa$(7)$ & 66.31 & 69.31 & 64.57 & 52.81  \\
    RoBERTa$(7)$+Mahalanobis & {\bf 66.31} &{\bf 69.89} & {\bf 64.82} & {\bf 53.46}  \\
    \specialrule{0em}{1pt}{1pt}
    \cdashline{1-5}
    \specialrule{0em}{1pt}{1pt}
    \multicolumn{5}{c}{\bf $k$Folden}\vspace{1pt}\\
    CNN-init emb & 62.94  & 65.08 & 54.28 &  50.27 \\ 
    BiLSTM-init emb & 61.05 & 67.81 & 56.98 & 49.96 \\ 
    BERT & 65.45 & 68.14 & 61.11 & 51.79 \\
    RoBERTa & {66.72} & 69.70 & 64.74 & 53.62 \\
    RoBERTa+Scaling & {66.72}  & 70.03 & 65.39  & 53.98  \\
    RoBERTa+Mahalanobis & {\bf 66.72} & {\bf 70.52}  & {\bf 65.81}  & {\bf 54.91}   \\\hline
    \multicolumn{5}{c}{\underline{\it AGNews-FL}}\vspace{1pt}\\
    \multicolumn{5}{c}{\bf Vanilla}\vspace{1pt}\\
    CNN-init emb & 80.55 & 62.94 & 52.70 & 30.54 \\ 
    BiLSTM-init emb & 81.36 & 63.71  & 54.77 & 31.90 \\ 
    BERT & 85.58 & 64.55 & 54.49 &  42.84   \\
    RoBERTa & 87.19 & 65.52  & 55.48 & 45.89  \\
    RoBERTa+Mahalanobis & 87.19 & 66.20 & {56.45}  &  46.95 \\
    RoBERTa-Dropout & {87.27} & 65.61 & 56.38 & 46.06  \\
    RoBERTa-Dropout+Mahalanobis & 87.27 & 66.53  &  57.11 & 46.89 \\
    RoBERTa$(4)$ & 87.55 & 65.81 & 56.89 & 46.19  \\
    RoBERTa$(4)$+Mahalanobis & {\bf 87.55} & {\bf 66.48}  & {\bf 57.49} & {\bf 46.92}   \\
    \specialrule{0em}{1pt}{1pt}
    \cdashline{1-5}
    \specialrule{0em}{1pt}{1pt}
    \multicolumn{5}{c}{\bf $k$Folden}\vspace{1pt}\\
    CNN-init emb & 82.21  & 63.45 & 53.98 &  34.71 \\ 
    BiLSTM-init emb & 84.33 & 64.44 & 55.01 & 35.68 \\ 
    BERT & 87.20 &  65.19 & 55.39 & 45.39 \\
    RoBERTa & 88.03 & 66.29  &  57.39 & 46.27 \\
    RoBERTa+Scaling & 88.03  & 66.84 & 58.07  & 46.75  \\
    RoBERTa+Mahalanobis & {\bf 88.03} & {\bf 66.89} & {\bf 58.26}  & {\bf 47.16}  \\\hline
    \multicolumn{5}{c}{\underline{\it AGNews-FM}}\vspace{1pt}\\
    \multicolumn{5}{c}{\bf Vanilla}\vspace{1pt}\\
    CNN-init emb & 79.81  & 79.63  & 53.50 & 54.72 \\ 
    BiLSTM-init emb & 82.51 & 79.46 & 52.86 & 55.33  \\ 
    BERT & 83.40  & 80.63  & 56.79 & 59.84    \\
    RoBERTa & 85.62 & 82.53 & 58.84 & 60.36  \\
    RoBERTa+Mahalanobis & 85.62 & {83.04} &  {59.96} &  62.26  \\
    RoBERTa-Dropout & {87.59} & 82.64 & 59.76 &  60.86  \\
    RoBERTa-Dropout+Mahalanobis & 87.59 & 83.14  & 59.23 & 61.88 \\
    RoBERTa$(4)$ & 88.16 & 82.85 & 60.44 & 61.95   \\
    RoBERTa$(4)$+Mahalanobis & {\bf 88.16} & {\bf 83.27} & {\bf 60.82} & {\bf 62.34}  \\
    \specialrule{0em}{1pt}{1pt}
    \cdashline{1-5}
    \specialrule{0em}{1pt}{1pt}
    \multicolumn{5}{c}{\bf $k$Folden}\vspace{1pt}\\
    CNN-init emb & 80.77 & 79.83 & 55.63 & 55.69  \\ 
    BiLSTM-init emb & 83.43 & 80.23 & 57.40 & 55.57 \\ 
    BERT & 84.55 & 81.35 & 58.19 & 62.89 \\
    RoBERTa & 88.92 & 83.61 & 60.88 & 63.42  \\
    RoBERTa+Scaling & 88.92 & 84.04 & 61.27  & 63.73   \\
    RoBERTa+Mahalanobis & {\bf 88.92} & {\bf 84.31} & {\bf 61.48} & {\bf 64.29}   \\\hline
    \multicolumn{5}{c}{\underline{\it Yahoo!Answers-FM}}\vspace{1pt}\\
    \multicolumn{5}{c}{\bf Vanilla}\vspace{1pt}\\
    CNN-init emb & 89.44 & 80.36  & 69.49 & 55.01 \\ 
    BiLSTM-init emb & 90.57  & 79.42 & 68.43 & 55.49  \\ 
    BERT & 93.25 & 82.71 & 74.55 & 57.82    \\
    RoBERTa & 94.73 & 83.81 & 76.47 & 58.62  \\
    RoBERTa+Mahalanobis & 94.73 & {84.51}  & {77.38}  & {59.86}  \\
    RoBERTa-Dropout &  {95.13} & 84.46  & 77.09 & 59.05  \\
    RoBERTa-Dropout+Mahalanobis  & { 95.13} & {84.90}  & {77.50} & {59.99} \\
    RoBERTa$(5)$ & 95.16  & 84.78 & 77.42 & 59.18  \\
    RoBERTa$(5)$+Mahalanobis & {\bf 95.16}  & {\bf 85.06} & {\bf 77.92} & {\bf 60.28}  \\
    \specialrule{0em}{1pt}{1pt}
    \cdashline{1-5}
    \specialrule{0em}{1pt}{1pt}
    \multicolumn{5}{c}{\bf $k$Folden}\vspace{1pt}\\
    CNN-init emb & 90.38  & 81.92 & 70.82 & 57.49  \\ 
    BiLSTM-init emb & 91.42 &  82.84 & 72.81 & 58.06 \\ 
    BERT & 94.74 &  84.15 &  76.92 & 58.34 \\
    RoBERTa & 95.56 & 85.50 & 78.52 & 59.10  \\
    RoBERTa+Scaling & 95.56 & 85.66 & 78.82  & 59.95  \\
    RoBERTa+Mahalanobis & {\bf 95.56}  & {\bf 85.83} & {\bf 78.88}  & {\bf 61.70}  \\\bottomrule
    \end{tabular}
    }
    \caption{Results of Semantic Shift (SS) datasets. The number in the bracket ($k$) denotes averaging $k$ model predictions and $k$ equals to the number of labels in the training dataset.}
    \label{exp:ss}
    \end{table}

\section{Benchmark Construction}
\label{benchmark}
Out-of-distribution data can be conceptually divided into two categories:  
non-semantic shift (NSS) and semantic shift (SS) and 
 datasets \cite{hsu2020generalized}. They are different in terms of whether a shift is related to the inclusion of new semantic categories: the training and 
 OOD 
 test examples in the NSS dataset come from different
 sub-categories of the same broader category.  
 For example, the training and OOD test sets in an NSS dataset 
 are both from the  
  ``car'' category, but  examples in the training set are able ``real car'', e.g. ``that's when they took out the fuel tank and poured it into a jug'', and all 
  OOD test 
   are about ``toy car'', e.g. ``Raleigh 2-year-old fills up toy car with 'gas' amidst shortage''. 
   For SS, 
   the training and 
 OOD 
 test examples in the SS dataset come from completely different categories.
 For example, 
 the training set contains labels ``car'' and ``bicycle'', and the test set has labels ``train'' and ``plane'', which have no intersections with training labels.
In this paper, we construct both SS and NSS text classification benchmarks for OOD detection. 

We construct  benchmarks on multi-class topic classification datasets. The topic classification task has less vocabulary overlap between ID and OOD data. 
We use data from 20NewsGroups \cite{joachims1996probabilistic}, Reuters-21578\footnote{\url{http://kdd.ics.uci.edu/databases/reuters21578/reuters21578.html}}, AG News \cite{del2005ranking} and Yahoo!Answers \cite{zhang2015character}. 
More details of the original datasets can be found at Appendix \ref{sec:appendixdata}.
The statistics of the benchmark are present in Table \ref{tab:benchmark}. 

We construct  NSS benchmarks as follows: 
\paragraph{20Newsgroups-6S}
This dataset is a modified version of 20Newsgroups. The original 20Newsgroups dataset has 20 newsgroups and each newsgroup (e.g., "comp.sys.ibm.pc.hardware") has a root subject topic (e.g., "comp"). We divide articles by its root subject and obtain 6 newsgroups ("comp", "rec", "sci", "religion", "politics" and "misc"). In this way, train and test data  share the same root topic labeled 
but have different fine-grained topic labels. 
The training and ID test data are from 11 sub-classes in 20News, while OOD test data are from the rest 9 sub-classes. 

\paragraph{AGNews-EXT}
This dataset is adapted from AG News and additional articles
come
 from the AG Corpus. The original AG News dataset has 4 classes ("World", "Sports", "Business", "Sci/Tech"). The training and ID test data in AGNews-EXT come from the 4 class labels in AG News, and the OOD test data are from the AG Corpus but have the same class labels as in AG News.
  
\paragraph{Yahoo-AGNews-five}
This dataset contains a subset of Yahoo!Answers and a subset of AG Corpus. The original Yahoo!Answers dataset has 10 classes, and we use 5 of them ("Health", "Science \& Mathematics", "Sports", "Entertainment \& Music", "Business \& Finance") for the training and ID test data. The OOD  test data are  selected from the 5 classes ("Health", "Sci/Tech", "Sports", "Entertainment", "Business") in AG Corpus.

We construct SS benchmarks as follows: 
\paragraph{Reuters-$m$K-$n$L}
This dataset is a modified version of Reuters. We first follow previous works \cite{yang1999re,joachims1998text} to use the ModApte split\footnote{\url{http://kdd.ics.uci.edu/databases/reuters21578/README.txt}} to remove documents belonging to multiple classes, and then considered only 10 classes ("Acquisitions", "Corn", "Crude", "Earn", "Grain", "Interest", "Money-fx", "Ship", "Trade" and "Wheat") with the highest numbers of training examples. The resulting dataset is called Reuters-ModApte. We train the model on a subset of Reuters-ModApte and test on the rest subset. Specifically, we train with $m$ topic articles and test the model on the other $n=10-m$ topics. In this paper, we use five settings: $(m,n)=(9,1)/(6,4)/(5,5)/(3,7)/(2,8)$.

\paragraph{AGNews-FL}
The dataset is adapted from AGNews and additional articles come from AG Corpus. In this setting, the training and ID test data are from the 4 classes ("World", "Sports", "Business", "Sci/Tech") in AGNews, and the OOD test data are from another 4 classes ("U.S.", "Europe", "Italia", "Software and Development") in AG Corpus.

\paragraph{AGNews-FM}
This dataset is adapted from AGNew and additional articles 
are taken
from the AG Corpus. In this setting, the training and ID data are from the 4 classes ("World", "Sports", "Business", "Sci/Tech") in AGNews, and the OOD test data are from another 4 classes ("Entertainment", "Health", "Top Stories", "Music Feeds") in AG Corpus. This dataset is easier than AGNews-FL because the OOD labels are more distinct from the ID labels regarding the label semantics.

\paragraph{Yahoo!Answers-FM}
This dataset is modified from the Yahoo!Answers dataset. We use five topic articles ("Health", "Science \& Mathematics", "Sports", "Entertainment \& Music", "Business \& Finance") for the training and ID tet data and use the other five unseen topics ("Society \& Culture", "Education \& Reference", "Computers \& Internet", "Family \& Relationships",  "Politics \& Government") for the OOD test data.

\section{Experiments}
\subsection{Experimental Setups}
We use both contextual and non-contextual model skeletons for experiments.
We use CNN and BiLSTM as the non-contextual model backbones. 
We follow the CNN-non-static model \cite{kim-2014-convolutional} as the CNN implementation and the BiLSTM model is of a single layer. Both CNN and BiLSTM have 300$d$ word vectors pretrained on Wikipedia 2014 using Glove \cite{glove2014}.
The average of the hidden states of all words is used as the feature for classification.  
We trained the non-contextual models with a batch size of 32 and an initial learning rate of 0.001 using the Adam \cite{kingma2014adam}. 
For contextual models, we use the officially pretrained BERT-uncased-base \cite{devlin2018bert} and RoBERTa-uncased-Base \cite{yinhan2019roberta} for comparison. 
We use AdamW \footnote{https://github.com/huggingface/transformers} to optimize all contextual models, with 0.01 weight decay and 1000 warmup steps. The learning rate was choosen in the range of $\{1e-5, 2e-5, 3e-5\}$. 
We use batch size in the range of $\{16, 24, 32\}$ for all experiments. And use dropout 0.2 for BERT and RoBERTa experiments. 

\subsection{Baselines}
We choose the following OOD detection methods for comparison: 

{\bf MSP}: The Maximum Softmax Probability method proposed by \citet{hendrycks2016baseline}. It uses the maximum probability in the final probability distribution over labels as the prediction score. If the maximum probability is under some specified threshold $\varphi\in[0,1]$, then the example would be classified as OOD. We tune the threshold on the dev set. This is the default setting for all model backbones.

{\bf Scaling}: The temperature scaling \cite{guo2017calibration} method leverages a temperature $T>0$ to sharpen or widen the probability distribution, and then treats the maximum probability as the final score. The temperature $T$ is chosen from $\{1, 10, 100, 1000,5000\}$ and is selected on the OOD validation set.

{\bf Mahalanobis}: \citet{lee2018simple} defined the confidence score using the Mahalanobis distance of a test example $\bm{x}$ with respect to the closest class-conditional distribution, which can be expressed as:
$\text{score}(\bm{x})=\min\limits_{c}(\psi(\bm{x})-\bm{\mu}_c)^\top\bm{\Sigma}^{-1}(\psi(\bm{x})-\bm{\mu}_c)$, where $\psi(\bm{x})$ is the vector representation of the input $\bm{x}$, $\bm{\mu}_c=\frac{1}{N_c}\sum\limits_{\bm{x}\in\mathcal{D}^{c}} \psi(\bm{x})$ is the centroid for class $c$ in the valid set $\mathcal{D}^\text{valid}$ and $\bm{\Sigma}=\frac{1}{N} \sum \limits_{c} \sum \limits_{\bm{x}\in\mathcal{D}^{c}} (\psi(\bm{x}) - \bm{\mu}_c) (\psi(\bm{x})-\bm{\mu}_c)^\top$ is the co-variance matrix. $N_c$ is the number of instances belongs to class $c$ in $\mathcal{D}^\text{valid}$. \\

{\bf Dropout}: \citet{gal2016dropout} casted dropout training as Bayesian inference for neural networks and obtained multiple predictions by running the model multiple times with dropout opened for a fixed input. These predictions are then averaged, giving the final probability distribution. Note that we can combine this method with the above three approaches.

More details regarding hyperparameter selection are present in Table \ref{tab:hyperparameter}.
Since the proposed strategy  uses the ensemble of K models, we also  implement
an ensemble of $k$ vanilla models. 

\begin{table}[t]
    \centering
    \scalebox{0.85}{
    \begin{tabular}{p{3cm}p{5cm}}\toprule
       {\bf Hyperparameter}  & {\bf Values to select} \\\midrule
        batch size &  \{16, 24, 32, 48\} \\
        dropout &  \{0.1, 0.2, 0.3\}\\
        weight decay &  \{0, 0.01\}\\
        max epochs &  \{3, 5, 8\}\\
        warmup ratio &  \{0, 0.1, 0.05\}\\
        learning rate &  \{1e-5, 2e-5, 3e-5\}\\
        learning rate decay & linear \\
        gradient clip & 1.0 \\
        {\bf MSP} threshold $\varphi$ & \{0, 0.001, 0.01, 0.05, 0.1, 0.2\} \\
        {\bf Scaling} temperature $T$ & \{1, 10, 100, 1000, 5000 \} \\
        {\bf Scaling} threshold $\varphi$ & \{0, 0.0005, 0.001, 0.0015, 0.002, 0.005, 0.01, 0.05, 0.1, 0.2\} \\
        number of passes in {\bf Dropout} & \{5, 10, 15, 20, 30\}  \\
        \bottomrule
    \end{tabular}
    }
    \caption{The range of hyperparameter values.}
    \label{tab:hyperparameter}
\end{table}

\subsection{Metrics}
\label{metrics}

We use accuracy (ACC) to evaluate model performances on the in-distribution testset and follow previous works \citep{hendrycks2016baseline, hsu2020generalized, lee2018simple} to employ three metrics for the OOD detection task, including AUROC, AUPR$_{out}$, TNR@95TPR.\

{\bf AUROC}: The AUROC is short for area under the receiver operating characteristic curve. The ROC curve is a graph plotting true negative rate against the false positive rate = FP/(FP+TN) by varying a threshold. 
This score is a threshold-independent evaluation metric and can be interpreted as the probability that a positive example has a greater detector score/value than a negative example \cite{tomroc06}. A random classifier has an AUROC score of 50\%. A higher AUROC value indicates a better OOD detection performance.

{\bf AUPR$_{out}$}: The AUPR is short for the area under the precision-recall curve. The precison-recall curve is a graph plotting the precision=TP/(TP+FP) against recall=TP/(TP+FN) by varying a threshold. AUPR$_{out}$ requires taking out-of-distribution data as the positive class. It is more suitable for highly imbalanced data compared to AUROC.

{\bf TNR@95TPR}: The TNR@95TPR is short for true negative rate (TNR) at 95\% true positive rate (TPR). 
The TNR@95TPR measures the true negative rate (TNR = TN/(FP+TN)) when the true positive rate (TPR = TP/(TP+FN)) is 95\%, where TP, TN, FP and FN denotes true positive, true negative, false positive and false negative, respectively. It can be interpreted as the probability that an example predicted incorrectly is misclassified as a corrected prediction when TPR is equal to 95\%.

\subsection{Results}
Experimental results for non-semantic shift and semantic shift benchmarks are shown in Table \ref{exp:nss} and Table \ref{exp:ss}, respectively. 
The first observation is that 
contextual models (BERT and RoBERTa) can achieve significantly better performances on both in-distribution and out-of-distribution datasets than non-contextual models (e.g., CNN, LSTM).
The second observation is that existing methods including Scaling, Mahalanobis and Dropout can improve ID and OOD performances.
The proposed $k$Folden framework introduces performance boost over 
the ensemble of its corresponding 
vanilla model (e.g., CNN, LSTM, Bert and RoBerta) in both ID and OOD evaluations. 
Additionally, we also find that 
$k$Folden is a flexible and general framework, which can be combined to existing OOD detection methods such as Mahalanobis, scaling and dropout, and can introduce addition  performance boosts in OOD detection. 

It is interesting to see that the improvements on SS datasets are greater than on NSS datasets when augmenting with the $k$Folden framework. This is because compared to NSS tasks, SS poses more variability in data distributions and requires a better generality from ID to OOD samples. $k$Folden serves this purpose well since it performs in a way as OOD simulation during training, which naturally addresses ID classification and OOD detection at the same time during training. This training paradigm wins better results for $k$Folden on SS data.


\subsection{The Ratio of Unseen Labels}
In this subsection, we explore the effect of unseen categories at different ratios. 
We use RoBERTa as the model backbone and conduct experiments on Reuters-$m$K-$n$L datasets, including $9$K-$1$L, $6$K-$4$L, $5$K-$5$L, $3$K-$7$L and $2$K-$8$L. 
We use accuracy and the error rate as evaluation metrics. The error rate represents the the proportion of OOD examples that are incorrectly classified to an in-distribution label, i.e., the maximum class probability is above the threshold tuned on the valid set.
Experimental results are shown in Table \ref{exp:dis-reuters}. 
As we can see from Table \ref{exp:dis-reuters}, the overall trend is that the error rate increases as more unseen text categories are added to the out-of-distribution test set.
Regarding specific models, we find that $k$Folden always outperforms Dropout, and the combination of $k$Folden and Mahalanobis leads to the best performance.
We speculate that this is because unlike Dropout which relies on the masking patterns within the neural network, the $k$Folden framework straightforwardly performs at the output, or the training objective level using the training data. This gives a direct learning signal for the model to learn to distinguish OOD examples.

\begin{table}[!t]
  \centering
  \scalebox{0.9}{
  \begin{tabular}{lcc}\toprule
    {\bf Model}& {\bf AUPR$\uparrow$}  & {\bf Error Rate$\downarrow$}\\\midrule
    \multicolumn{3}{c}{\underline{\it Reuters-9K-1L}}\vspace{1pt}\\
    RoBERTa & 79.77 & 36.61  \\
    RoBERTa+Dropout & 80.07 & 32.74   \\
    $k$Folden RoBERTa & 81.53 & 30.63  \\
    $k$Folden RoBERTa+mahal & {\bf 81.68} & {\bf 29.75} \\
    \specialrule{0em}{1pt}{1pt}
    \cdashline{1-3}
    \specialrule{0em}{1pt}{1pt}
    \multicolumn{3}{c}{\underline{\it Reuters-6K-4L}}\vspace{1pt}\\
    RoBERTa & 78.52 & 36.26   \\
    RoBERTa+Dropout & 79.73 & 36.13 \\
    $k$Folden RoBERTa & 80.83 & 35.76  \\
    $k$Folden RoBERTa+mahal & {\bf 82.74} & {\bf 35.49} \\
    \specialrule{0em}{1pt}{1pt}
    \cdashline{1-3}
    \specialrule{0em}{1pt}{1pt}
    \multicolumn{3}{c}{\underline{\it Reuters-5K-5L}}\vspace{1pt}\\
    RoBERTa & 89.56 & 42.83 \\
    RoBERTa+Dropout & 90.25  & 41.36  \\
    $k$Folden RoBERTa & 91.76 & 40.99  \\
    $k$Folden RoBERTa+mahal & {\bf 92.08} & {\bf 40.76} \\
    \specialrule{0em}{1pt}{1pt}
    \cdashline{1-3}
    \specialrule{0em}{1pt}{1pt}
    \multicolumn{3}{c}{\underline{\it Reuters-3K-7L}}\vspace{1pt}\\
    RoBERTa & 95.64 & 46.27  \\
    RoBERTa+Dropout & 96.14 & 45.89  \\
    $k$Folden RoBERTa & 96.75 & 44.82  \\
    $k$Folden RoBERTa+mahal & {\bf 96.83} & {\bf 43.69}   \\
    \specialrule{0em}{1pt}{1pt}
    \cdashline{1-3}
    \specialrule{0em}{1pt}{1pt}
    \multicolumn{3}{c}{\underline{\it Reuters-2K-8L}}\vspace{1pt}\\
    RoBERTa & 97.35 & 58.14   \\
    RoBERTa+Dropout & 97.56  & 57.62 \\
    $k$Folden RoBERTa & 97.83 & 56.80   \\
    $k$Folden RoBERTa+mahal & {\bf 97.91} & {\bf 56.06}  \\\bottomrule
    \end{tabular}
    }
    \caption{Results on Reuters-$m$K-$n$L OOD test sets. The Reuters dataset contains 10 label categories. We use $m$ to represent the number of labels in ID training set and $n$ for the number of categories in OOD testset.}
    \label{exp:dis-reuters}
\end{table}

\section{Conclusion}
In this paper, we propose a simple yet effective framework $k$Folden for OOD detection. It works by mimicking the behaviors of detecting out-of-distribution examples during training without the use of any external data. 
We also develop a benchmark on top of existing widely used datasets for text classification OOD detection. This benchmark contains both semantic shift and non-semantic shift data, which would benefit a comprehensive examination to the ability of OOD detection methods.  
Through experiments and analyses, we show that the proposed $k$Folden framework outperforms strong OOD detection baselines on the constructed benchmark, and combining $k$Folden and other post-hoc methods leads to the most performance gains. 
We hope the proposed method and the created benchmark can facilitate further researches in related areas.

\section*{Acknowledgement}
This work was supported by Key-Area Research and Development Program of Guangdong Province (No. 2019B121204008).
We thank the High-Performance Computing Platform at Peking University and the PCNL Cloud Brain for providing platforms of data analysis and model training.

\bibliography{custom}
\bibliographystyle{acl_natbib}

\appendix

\section{Dataset Details}
\label{sec:appendixdata}
\subsection{Original Datasets}
In this paper, We use data from 20NewsGroups \cite{joachims1996probabilistic}, Reuters-21578\footnote{\url{http://kdd.ics.uci.edu/databases/reuters21578/reuters21578.html}}, AG News \cite{del2005ranking} and Yahoo!Answers \cite{zhang2015character} to construct our evaluation benchmark. Details regarding these four datasets are present below:
\begin{itemize}[noitemsep,topsep=0pt,parsep=0pt,partopsep=0pt]
  \item {\bf 20Newsgroups}\footnote{\url{https://kdd.ics.uci.edu/databases/20newsgroups/20newsgroups.html}}: 20Newsgroups is a collection of approximate 20,000 newsgroup documents, partitioned (nearly) evenly across 20 different newsgroups. Each newsgroup corresponds to a different topic. Some of the newsgroups are very closely related to each other (e.g., ``comp.sys.pc.hardware'' and ``comp.sys.mac.hardware''), while others are highly unrelated (e.g., ``misc.forsale'' and ``soc.religion.christian'').
  \item {\bf AG News}\footnote{\url{http://groups.di.unipi.it/~gulli/AG\_corpus\_of\_news\_articles.html}}: AG News is a subdataset of AG's corpus of news articles constructed by assembling titles and description fields of articles from the four largest classes ("World", "Sports", "Business", "Sci/Tech") of AG Corpus. AG News contains 30,000 training and 1,900 test samples per class. 
  \item {\bf Yahoo!Answers}\footnote{\url{https://drive.google.com/drive/folders/0Bz8a\_Dbh9Qhbfll6bVpmNUtUcFdjYmF2SEpmZUZUcVNiMUw1TWN6RDV3a0JHT3kxLVhVR2M}}: Yahoo!answers was constructed by \citet{zhang2015character} and composed of 10 largest main categories from Yahoo!Answers Comprehensive Questions and the Answers version 1.0 dataset. Each class contains 140,000 training samples and 5,000 testing samples. Labels in the dataset include "Society \& Culture", "Science \& Mathematics", "Health", "Education \& Reference", "Computers \& Internet", "Sports", "Business \& Finance", "Entertainment \& Music", "Family \& Relationships", and "Politics \& Government". 
  \item {\bf Reuters-21578}\footnote{\url{http://kdd.ics.uci.edu/databases/reuters21578/reuters21578.html}}: Reuters-21578 is a collection of 10,788 documents from the Reuters financial newswire service, partitioned into a training set with 7,769 documents and a test set with 3,019 documents. The distribution of categories in the Reuters-21578 corpus is highly skewed, with 36.7\% of the documents in the most common category, and only 0.0185\% (2 documents) in each of the five least common categories. There are 90 categories in the corpus. Each document belongs to one or more categories. The average number of categories per document is 1.235, and the average number of documents per category is about 148, or 1.37\% of the corpus.
\end{itemize}

\subsection{Benchmark Construction}
We construct our NSS benchmarks as follows: 
\begin{table}[t]
    \center
    \scalebox{0.7}{
    \begin{tabular}{lcc}\toprule
    {\bf Label} &  {\bf Train\&ID-X} & {\bf OOD-X}  \\\midrule
    {\bf comp} & comp.graphics & comp.sys.mac.hardware   \\
    & comp.sys.ibm.pc.hardware & comp.windows.x  \\
    & comp.os.ms-windows.misc &     \\ \hline
    {\bf rec} & rec.autos & rec.sport.baseball   \\
    & rec.motorcycles & rec.sport.hockey  \\ \hline
    {\bf sci} & sci.crypt & sci.med   \\
    & sci.electronics & sci.space \\ \hline
    {\bf religion} & talk.religion.misc & alt.atheism   \\
    &  & soc.religion.christian  \\\hline
    {\bf politics}& talk.politics.guns & talk.politics.mideast   \\
    & talk.politics.misc &   \\ \hline
    {\bf misc} & misc.forsale &   \\
    \bottomrule
    \end{tabular}
    }
    \caption{Merging labels from 20News for 20News-6S.}
    \label{benchmark:20news6s}
\end{table}
\paragraph{20Newsgroups-6S}
This dataset is a modified version of 20Newsgroups. The original 20Newsgroups dataset has 20 newsgroups and each newsgroup (e.g., "comp.sys.ibm.pc.hardware") has a root subject topic (e.g., "comp"). We divide articles by its root subject and obtain 6 newsgroups ("comp", "rec", "sci", "religion", "politics" and "misc"). For example, the original label "comp.sys.ibm.pc.hardware" becomes "comp". Hence, train and test data share the same labels but may come from different data distributions. Data in the five sets do not overlap. We show the used classes for each of the following sets in Table \ref{benchmark:20news6s}.
\begin{itemize}[noitemsep,topsep=0pt,parsep=0pt,partopsep=0pt]
  \item {\bf TrainingSet} We use 8,283 articles from the trainset in 20Newsgroups belonging to 11 sub-classes. Each class contains 753 articles.
  \item {\bf ID-ValidSet} We use 1,034 articles from the trainset in 20Newsgroups belonging to 11 sub-classes. Each class contains 94 articles.
  \item {\bf ID-TestSet} We use 1,034 articles from the testset in 20Newsgroups belonging to 11 sub-classes. Each class contains 94 articles.
  \item {\bf OOD-ValidSet} We use 846 articles from the trainset in 20Newsgroups belonging to the other 9 sub-classes in 20Newsgroups. Each class contains 94 articles.
  \item {\bf OOD-TestSet} We use 846 articles from the testset in 20Newsgroups belonging to the other 9 sub-classes in 20Newsgroups. Each class contains 94 articles.
\end{itemize}

\paragraph{AGNews-EXT}
This dataset contains data from AG-News and additional articles from AG Corpus. In this setting, the training and ID data are from the same 4 labels ("World", "Sports", "Business", "Sci/Tech"). OOD data are from the same 4 labels but use articles in AG Corpus intead of AG-News. Data in the five sets do not overlap.
\begin{itemize}[noitemsep,topsep=0pt,parsep=0pt,partopsep=0pt]
  \item {\bf TrainingSet} We use 112,400 articles from the trainset in AG-News with 4 classes. Each class contains 28,100 articles.
  \item {\bf ID-ValidSet} We use 7,600 articles from the trainset in AG-News with the same 4 classes as TrainingSet. Each class has 1,900 articles.
  \item {\bf ID-TestSet} We use 7,600 articles from the testset in AG-News with the same 4 classes as TrainingSet. Each class has 1,900 articles. 
  \item {\bf OOD-ValidSet} We assemble titles and description fields of articles in AG Corpus from the same 4 classes as TrainingSet. Each class has 1,900 articles. 
  \item {\bf OOD-TestSet} We assemble titles and description fields of articles in AG Corpus from the same 4 classes as TrainingSet. Each class has 1,900 articles. 
\end{itemize}

\paragraph{Yahoo-AGNews-five}
This dataset contains a subset of Yahoo!Answers and a subset of AG Corpus. The original Yahoo!Answers dataset has 10 classes, and we use 5 of them ("Health", "Science \& Mathematics", "Sports", "Entertainment \& Music", "Business \& Finance") for the training and ID data. The OOD data are from the 5 classes ("Health", "Sci/Tech", "Sports", "Entertainment", "Business") in AG Corpus. Data in the five sets do not overlap.
\begin{itemize}[noitemsep,topsep=0pt,parsep=0pt,partopsep=0pt]
  \item {\bf TrainingSet} We use 675,000 articles from the trainset in Yahoo!Answers with 5 classes. Each class contains 135,000 articles. 
  \item {\bf ID-ValidSet} We use 25,000 articles from the trainset in Yahoo!Answers with the same 5 classes as TrainingSet. Each class contains 5,000 articles. 
  \item {\bf ID-TestSet} We use 25,000 articles from the testset in Yahoo!Answers with the same 5 classes as TrainingSet. Each class contains 5,000 articles.
  \item {\bf OOD-ValidSet} We assemble titles and description fields of articles in AG Corpus from the same 5 classes as TrainingSet. Each class contains 5,000 articles.
  \item {\bf OOD-TestSet} We assemble titles and description fields of articles in AG Corpus from the same 5 classes as TrainingSet. Each class contains 5,000 articles.
\end{itemize}

We construct SS benchmarks as follows: 
\paragraph{Reuters-$m$K-$n$L}
\begin{table*}[t]
    \centering
    \scalebox{0.90}{
    \begin{tabular}{lcccccccccc}\toprule
    {\bf Train } &  {\bf Acq} & {\bf Corn} & {\bf Crude} &  {\bf Earn} & {\bf Grain} & {\bf Interest} & {\bf Money-fx} & {\bf Ship} & {\bf Trade} & {\bf Wheat} \\\midrule
    {\bf Reuters} & 1615 & 175 & 383 & 2817 & 422 & 343 & 518 & 187 & 356 & 206   \\\hline
    {\bf Reuters-$9$K-$1$L} & 1615 & 175 & 383 & 2817 & 422 & 343 & 518 & N/A  & 356 & 206   \\\hline
    {\bf Reuters-$6$K-$4$L} & 1615 & 175 & N/A & 2817 & 422 & 343 & 518 & N/A  & N/A & N/A   \\\hline
    {\bf Reuters-$5$K-$5$L} & 1615 & N/A  & N/A  & 2817 & 422 & 343 & 518 & N/A  & N/A  & N/A    \\\hline
    {\bf Reuters-$3$K-$7$L} & 1615 & N/A  & N/A  & 2817 & N/A & N/A  & 518 & N/A  & N/A  & N/A    \\\hline
    {\bf Reuters-$2$K-$8$L} & 1615 & N/A  & N/A  & 2817 &N/A  & N/A  & N/A  & N/A  & N/A  & N/A    \\\hline
    \bottomrule
    {\bf ID Test} &  {\bf Acq} & {\bf Corn} & {\bf Crude} &  {\bf Earn} & {\bf Grain} & {\bf Interest} & {\bf Money-fx} & {\bf Ship} & {\bf Trade} & {\bf Wheat} \\\midrule
    {\bf Reuters} & 719 & 56 & 189 & 1087 & 149 & 131 & 179 & 89 & 117 & 71   \\\hline
    {\bf Reuters-$9$K-$1$L} & 719 & 56 & 189 & 1087 & 149 & 131 & 179& N/A & 117 & 71 \\\hline
    {\bf Reuters-$6$K-$4$L} & 719 & 56  & N/A & 1087 & 149 & 131 & 179 & N/A & N/A & N/A   \\\hline
    {\bf Reuters-$5$K-$5$L} & 719   & N/A & N/A & 1087  & 149 & 131 & 179 &  N/A & N/A  & N/A   \\\hline
    {\bf Reuters-$3$K-$7$L} & 719  & N/A & N/A & 1087 & N/A & N/A & 179  & N/A & N/A & N/A   \\\hline
    {\bf Reuters-$2$K-$8$L} & 719  &  N/A &  N/A & 1087 &  N/A &  N/A &  N/A &  N/A &  N/A &  N/A   \\ \bottomrule
     {\bf OOD Test} &  {\bf Acq} & {\bf Corn} & {\bf Crude} &  {\bf Earn} & {\bf Grain} & {\bf Interest} & {\bf Money-fx} & {\bf Ship} & {\bf Trade} & {\bf Wheat} \\\midrule
    {\bf Reuters} & 719 & 56 & 189 & 1087 & 149 & 131 & 179 & 89 & 117 & 71   \\\hline
    {\bf Reuters-$9$K-$1$L} & N/A  & N/A  & N/A  & N/A  & N/A  & N/A  & N/A  & 89 & N/A  & N/A    \\\hline
    {\bf Reuters-$6$K-$4$L} & N/A  & N/A  & 189 & N/A  & N/A  & N/A  & N/A  & 89 & 117 & 71   \\\hline
    {\bf Reuters-$5$K-$5$L} & N/A  & 56 & 189 & N/A  & 149 & 131 & 179 & N/A  & N/A  & N/A    \\\hline
    {\bf Reuters-$3$K-$7$L} & N/A  & 56 & 189 & N/A  & 149 & 131 & N/A  & 89 & 117 & 71   \\\hline
    {\bf Reuters-$2$K-$8$L} & N/A  & 56 & 189 & N/A & 149 & 131 & 179 & 89 & 117 & 71   \\\bottomrule
    \end{tabular}
    }
    \caption{Data statistics for Reuters-$m$K-$n$K datasets.}
    \label{sec:appendixreuters}
\end{table*}
This dataset is a modified version of Reuters. We first follow previous works \cite{yang1999re,joachims1998text} to use the ModApte split\footnote{\url{http://kdd.ics.uci.edu/databases/reuters21578/README.txt}} to remove documents belonging to multiple classes, and then considered only 10 classes ("Acquisitions", "Corn", "Crude", "Earn", "Grain", "Interest", "Money-fx", "Ship", "Trade" and "Wheat") with the highest numbers of training examples. The resulting dataset is called Reuters-ModApte. We train the model on a subset of Reuters-ModApte and test on the rest subset. Specifically, we train with $m$ topic articles and test the model on the other $n$ ($n=10-m$) topic articles. In this paper, we use five settings: Reuters-9K-1L, Reuters-6K-4L, Reuters-5K-5L, Reuters-3K-7L and Reuters-2K-8L. This task is difficult because the resulting datasets are highly unbalanced. All documents in train/valid/test come from Reuters-21578. Data in the five sets do not overlap. Data statistics can be found in Table \ref{sec:appendixreuters}.
\begin{itemize}[noitemsep,topsep=0pt,parsep=0pt,partopsep=0pt]
  \item {\bf TrainingSet} We choose articles in the trianset of Reuters-ModApte belonging to $m$ topics.  
  \item {\bf ID-ValidSet} We choose articles in the valid set of Reuters-ModApte belonging to $m$ topics.
  \item {\bf ID-TestSet} We choose articles in the test set of Reuters-ModApte belonging to $m$ topics.
  \item {\bf OOD-ValidSet} We choose articles in the valid set of Reuters-ModApte belonging to $n$ topics.
  \item {\bf OOD-TestSet} We choose articles in the test set of Reuters-ModApte belonging to $n$ topics.
\end{itemize}

\paragraph{AGNews-FL}
The dataset is composed of data from AGNews and additional articles from the AG Corpus. In this setting, the training and ID data are from the 4 classes ("World", "Sports", "Business", "Sci/Tech") in AGNews, and the OOD data are from another 4 classes ("U.S.", "Europe", "Italia", "Software and Development") in AG Corpus. It is noteworthy that these two sets of labels are similar in semantics, e.g., "U.S." to "World", "Europe" to "Sports" and "Software and Development" to "Sci/Tech". This makes the task more challenging than AGNews-FM, which will be introduced below.
Data in the five sets do not overlap.
\begin{itemize}[noitemsep,topsep=0pt,parsep=0pt,partopsep=0pt]
  \item {\bf TrainingSet} We use 116,000 articles from the trainset in AG-News belonging to 4 classes. Each class contains 29,000 articles. 
  \item {\bf ID-ValidSet} We use 4,000 articles from the trainset in AG-News. Each class has 1,000 articles.
  \item {\bf ID-TestSet} We use 4,000 articles from the testset in AG-News. Each class has 1,000 articles.
  \item {\bf OOD-ValidSet} We assemble titles and description fields of articles in AG Corpus from another 4 classes different from AG-News. There are 4,000 articles and 1,000 articles per class.
  \item {\bf OOD-TestSet} We assemble titles and description fields of articles in AG Corpus from another 4 classes different from AG-News. There are 4,000 articles and 1,000 articles per class.
\end{itemize}

\paragraph{AGNews-FM}
The dataset is composed of data from AGNews and additional articles from the AG Corpus. In this setting, the training and ID data are from the 4 classes ("World", "Sports", "Business", "Sci/Tech") in AGNews, and the OOD data are from another 4 classes ("Entertainment", "Health", "Top Stories", "Music Feeds") in AG Corpus. This dataset is easier than AGNews-FL because the OOD labels are more distinct from the ID labels regarding the label semantics. Data in the five sets do not overlap.
\begin{itemize}[noitemsep,topsep=0pt,parsep=0pt,partopsep=0pt]
  \item {\bf TrainingSet} We use 116,000 articles from the trainset in AG-News belonging to 4 classes. Each class contains 29,000 articles. 
  \item {\bf ID-ValidSet} We use 4,000 articles from the trainset in AG-News. Each class has 1,000 articles.
  \item {\bf ID-TestSet} We use 4,000 articles from the testset in AG-News. Each class has 1,000 articles.
  \item {\bf OOD-ValidSet} We assemble titles and description fields of articles in AG Corpus from another 4 classes different from AG-News. There are 4,000 articles and 1,000 articles per class.
  \item {\bf OOD-TestSet} We assemble titles and description fields of articles in AG Corpus from another 4 classes different from AG-News. There are 4,000 articles and 1,000 articles per class.
\end{itemize}

\paragraph{Yahoo!Answers-FM}
This dataset is modified from the Yahoo!Answers dataset. We use five topic articles ("Health", "Science \& Mathematics", "Sports", "Entertainment \& Music", "Business \& Finance") for the training and ID data and use the other five unseen topics ("Society \& Culture", "Education \& Reference", "Computers \& Internet", "Family \& Relationships",  "Politics \& Government") for the OOD data. Data in the five sets do not overlap.
\begin{itemize}[noitemsep,topsep=0pt,parsep=0pt,partopsep=0pt]
  \item {\bf TrainingSet} We use 680,000 examples belonging to five categories in Yahoo!Answers, 136,000 samples per class.  
  \item {\bf ID-ValidSet} We use 20,000 examples belonging to five categories in Yahoo!Answers. 4,000 samples per class. 
  \item {\bf ID-TestSet} We use 25,000 examples belonging to five categories in Yahoo!Answers. 5,000 samples per class. 
  \item {\bf OOD-ValidSet} The data are from another five categories in Yahoo!Answers. The OOD-ValidSet contains 20,000 articles with 4,000 per class.  
  \item {\bf OOD-TestSet} The data are from another five categories in Yahoo!Answers. The OOD-TestSet contains 25,000 articles with 5,000 per class. 
\end{itemize} 


\end{document}